\documentclass[runningheads]{llncs}

\usepackage{microtype}
\usepackage{subfigure}
\usepackage{booktabs} 
\usepackage{amsmath}
\usepackage{mathtools}  
\usepackage{graphicx}
\usepackage[T1]{fontenc}
\usepackage{xcolor}
\usepackage{ulem}

\usepackage{comment}
\usepackage[utf8]{inputenc}
\usepackage{amsmath}
\usepackage{amssymb}

\DeclareUnicodeCharacter{0307}{\text{.}}

\begin{document}
%

\title{A Lightweight Measure of Classification Difficulty from Application Dataset Characteristics}
\titlerunning{Measuring Classification Difficulty}

%
\authorrunning{B. B. Cao et al.}
%

%
\author{Bryan Bo Cao\inst{1,2} \and
Abhinav Sharma\inst{1,2} \and
Lawrence O'Gorman\inst{1} \and
Michael Coss\inst{1} \and
Shubham Jain\inst{2}}
\institute{Nokia Bell Labs, Murray Hill, NJ, USA \and
Stony Brook University, Stony Brook, NY, USA\\
\email{\inst{1}\{bryan.cao,abhinav.6.sharma\}@nokia.com}\\
\email{\inst{1}\{larry.o\_gorman,mike.coss\}@nokia-bell-labs.com}\\
\email{\inst{2}\{boccao,abhinsharma,jain\}@cs.stonybrook.edu}}

\maketitle              
%
%
    Although accuracy and computation benchmarks are widely available to help choose among neural network models, these are usually trained on datasets with many classes, and do not give a good idea of performance for few ($<10$) classes. The conventional procedure to predict performance involves repeated training and testing on the different models and dataset variations. We propose an efficient cosine similarity-based classification difficulty measure $S$ that is calculated from the number of classes and intra- and inter-class similarity metrics of the dataset. After a single stage of training and testing per model family, relative performance for different datasets and models of the same family can be predicted by comparing difficulty measures – without further training and testing. Our proposed method is verified by extensive experiments on 8 CNN and ViT models and 7 datasets. Results show that $S$ is highly correlated to model accuracy with correlation coefficient $|r| = 0.796$, outperforming the baseline Euclidean distance at $|r| = 0.66$. We show how a practitioner can use this measure to help select an efficient model $6$ to $29\times$ faster than through repeated training and testing. We also describe using the measure for an industrial application in which options are identified to select a model 42\% smaller than the baseline YOLOv5-nano model, and if class merging from 3 to 2 classes meets requirements, 85\% smaller.





\section{Introduction}
\label{sec:Introduction}

Much information is available to compare neural network models. Besides inherent features such as size and speed, model performance is measured by accuracy on public datasets. This information is invaluable for comparing models, but it is often far-removed from predicting how a model will perform on a particular application. One reason for this is that most public datasets have many classes (e.g., 1000 for ImageNet \cite{deng2009imagenet}, 80 for COCO \cite{lin2014microsoft}). But many applications have far fewer classes. For example, 7 for object identification while driving \cite{applDriving2021}, 6 for wildlife animal detection \cite{applWildlifeAustralia2017}, 4 for cancer cell classification \cite{appl_cancer}, and 2 each for crowd \cite{appl_crowd}, cattle \cite{appl_cattle}, hardhat \cite{appl_hardhat} and ship, SAR detection \cite{appl_shipSAR}, to name but a few. Another reason is that the difficulty of the instances in the datasets is often unknown in both the benchmark and application dataset. For both these reasons, it is difficult to extrapolate from performance for the large public datasets to performance on a particular application of much different number of classes and similarity between classes.

The classification difficulty of an application depends in large part upon its dataset characteristics. For instance, an application whose classes have very similar features will generally have higher classification difficulty than for one with lower inter-class similarity. Although the general relationship between data characteristics and classification difficulty is known, there are benefits to quantifying this by means of a difficulty measure. For instance, knowing the classification performance of a model on a dataset, how will it perform on a different dataset? One could train and test, or alternatively one could compare dataset difficulties. We show how the latter is a lightweight approach requiring much less computation especially for few-class ($<10$ classes) applications.

In this paper, we propose a quantitative, lightweight measure of classification difficulty for an application dataset based upon the number of classes and class similarity. Although the measure can be applied to datasets of all sizes, it is most suitable for datasets of fewer classes (<10), which are typical of many practical applications. For these applications, the difficulty measure can help direct a practitioner to a model whose balance of accuracy and computational efficiency meet the requirements.  

Contributions of this work are summarized as follows:
\begin{enumerate}
\item 
\textbf{Analytical} -- Determination of a mathematical relationship between classification difficulty and the dataset characteristics.
\item 
\textbf{Experimental} -- Experimental results showing how dataset classification difficulty relates to model performance.
\item
\textbf{Practical} -- Quantifying the efficiency advantage of using dataset classification difficulty for smaller size and lower power model selection and, dataset modification.
\item 
\textbf{Use Case} -- An industrial example of how dataset classification difficulty is used to adjust application specifications toward a more efficient model choice.
\end{enumerate}

This paper is organized as follows. In Section \ref{sec:relatedWork}, we discuss related literature on model selection and difficulty measures. In Section \ref{sec:Method}, a new measure of dataset classification difficulty is presented. We present experimental evidence of the quantitative relationship between dataset characteristics and model performance in Section \ref{sec:Exp}. We show an example of how the measure is used in an industrial application in Section \ref{sec:Appl_robot}. The results are discussed and summarized in Section \ref{sec:Conclusion}.

\section{Related Work}
\label{sec:relatedWork}

\subsection{Model Selection}
\label{sec:modelSelection}
In recent years, a plethora of neural architectures have been designed, trained, and made easily available such that a practitioner will usually select a model rather than designing or training one from scratch. However, with the increasing number of efficient architectures (e.g. MobileNets \cite{howard2017mobilenets,sandler2018mobilenetv2,koonce2021mobilenetv3}~, SqueezeNet \cite{koonce2021squeezenet}~ ShuffleNets \cite{zhang2018shufflenet,ma2018shufflenet}~, EfficientNet \cite{tan2019efficientnet}, etc.), selecting a proper model to satisfy an application's requirements becomes even more challenging. To select a model, existing approaches can be categorized into four options: 1) off-the-shelf, 2) transfer learning plus targeted training, 3) scaled model selection, and 4) selection from model repository (with caveats as described below). 

For the first option, numerous off-the-shelf models \cite{pytorchhub,tfhub,hfhub}~ have been trained on a dataset containing the same classes as the application, for instance for pedestrian detection \cite{pedestrian2017}, flower classification \cite{flowers2018}, fish classification \cite{fish2018}, and food detection \cite{food2018}. Although this can save substantial time from data collection and training, it often fails in real-world applications due to a feature shift in deployed environments due to such factors as different camera capture angles, backgrounds, scales, etc.

A second option is transfer learning \cite{transferLearn2021} by which a model is trained on a larger, standard dataset such as ImageNet \cite{deng2009imagenet} or COCO \cite{lin2014microsoft}, stripped of its classification layer (leaving the backbone), then fine-tuned on objects of the application of interest. A drawback to this popular choice is that the backbone incorporates extraneous features than are often needed for the application classes. Although transfer learning facilitates the selection of a neural model with high accuracies, the model will inevitably be larger than one trained on only the classes of interest for the same level of accuracy.

A third option is to choose from a scaled model family -- a collection of models that share the same general architecture, but whose size (width and depth) are adjusted with a scaling factor. Examples include the EfficientNet family \cite{tan2019efficientnet}~ from B0 to B7, and YOLO family \cite{glenn_jocher_2021_5563715} from nano to extra-large scales. This paper focuses on efficiency for small, practical applications. Therefore, experiments on the smaller sizes of the EfficientNet image classifier and the YOLO object detector \cite{ganesh2022yolo} are chosen for testing our small model regime.

The fourth option is to select a pre-trained model from a model repository \cite{SommelierModelRepository_2022,zhou2023learnware}. This is a fast way to start using a model, but it is different than the focus of this paper in two ways. Our focus is on selecting efficient models to match dataset characteristics. The most efficient models must be trained on the dataset of the application. This is unlike a  model repository whose models may be efficient, but are unlikely for the application dataset specifically. The second difference is that our methods are directed to the \textit{data side} versus the \textit{model side} \cite{ZestForLIME_2022,modelDiff2023}. This enables different application datasets to be compared by their classification difficulty rather than different models to be compared by their performance on pre-trained datasets.

\subsection{Image Classification Difficulty}
No matter which type of classifier is used, the empirically observed behavior of classifiers is strongly data dependent. Previous to the widespread adoption of neural networks, classification difficulty was largely measured by the ability to distinguish classes volumetrically in multi-dimensional feature space. A classical measure is Fisher's Discriminant Ratio, by which a large difference in class means and small sum of their variances describes a less difficult classification problem \cite{dudaHart1973}. In \cite{ClassComplexityHo2002}, complexity measures are described that include feature overlap, feature efficiency, separability of classes, and geometry of the class manifolds. Although the embedding space of a neural network is also a feature space, neural networks often have much higher dimensionality of nonintuitive (machine-learned) features, which have the ability to better distinguish highly non-linear class boundaries, thus leading to neural network classification difficulty measures different from these previous measures.



A number of image difficulty measures have been proposed. For metric learning \cite{bellet2013survey,metricLearn2017,SimilarityMetricLearning2021}, the loss function is set to minimize the similarity between images of the same class during training, where similarity is measured as the dot product between embeddings (usually from the last hidden layer of the model) of two images \cite{Ionescu_2016_CVPR,medDiagnose_2022,AngularGap2022}. Another way to measure difficulty is by classification error on a difficulty-scaled range of datasets \cite{DatasetClassification_2021,imagenetDifficulty2022,DLLagree2022} or models \cite{ScaledInference_2022}. Machine difficulty scores can also be used to prune filters associated with easier features during training \cite{dynamicPrune_2023}, and by more highly weighting filters of difficult features during inference \cite{weightedModel2020}.

One difference from these previous papers is that we focus upon \textit{classification difficulty}. Many references calculate \textit{single-image} difficulty for purposes of curriculum, or simple-to-difficult, learning \cite{Ionescu_2016_CVPR,appalaraju2017image,medDiagnose_2022,AngularGap2022} and scaled model selection \cite{Ionescu_2Stage_2018}. In \cite{weightedModel2020}, intra-class difficulty is measured for the purpose of weighting classes differently during training. In contrast to single-image difficulty, we incorporate intra- and inter-class similarity in determining a difficulty measure for application datasets containing many images of multiple classes, as shown in Section \ref{sec:Method}. And is Section \ref{sec:Exp}, we show by experiment how the measure varies for different numbers of classes, similarities, models, and datasets.

\section{Dataset Classification Difficulty Measure}
\label{sec:Method}
\label{sec:method_difficMeasure}


Cosine similarity is a common measure used to quantify the similarity between two feature vectors, 
\begin{equation}
\footnotesize
    \textrm{S}(\textbf{x}_i, {\textbf{x}_j}) = 
    {\cos({\textbf{z}_i, \textbf{z}_j}) = \frac{\textbf{z}_i \cdot {\textbf{z}_j}}{||{\textbf{z}_i}|| \: ||{\textbf{z}_j}||}},
     \label{eqn:cosineSimilarity}
\end{equation}
where {$\textbf{z}_i$} and {$\textbf{z}_j$} are feature vectors {of image} $i$ and $j$ respectively.


Whereas equation \ref{eqn:cosineSimilarity} is the similarity between two vector instances, we are interested in the average similarity between pairs of instances in the same class, and pairs of instances between classes respectively,  

\begin{equation}
\footnotesize
    \textbf{intra-class: } S_R({C}) = \frac{1}{n_1}\sum_{i,j \in C, i \neq j }S(\textbf{x}_{i}, {\textbf{x}_{j}})
     \label{eqn:S1}
\end{equation}
\begin{equation}
\footnotesize
    \textbf{inter-class: } S_E({C_{a}, C_{b}}) = \frac{1}{n_2}\sum_{i \in {C_{a}}, j \in {C_{b}}}S(\textbf{x}_{i}, {\textbf{x}_{j}}) 
     \label{eqn:S2}
\end{equation}
where $n_1$ is the number of intra-class pair combinations from a single class set of instances {$\textbf{x} \in C$}, and $n_2$ is the number of inter-class pair combinations between instances of two classes {\textbf{$x_{i}$} $\in C_{a}$ and \textbf{$x_{j}$} $\in C_{b}$}.

For a classification problem with $N_{CL}$ classes, the average intra- and inter-class similarities are respectively,
\begin{equation}
\footnotesize
    \bar{S}_R = \frac{1}{{n_3}}\sum S_{R}, \hspace{0.5cm}
    \bar{S}_E = \frac{1}{{n_4}}\sum S_{E},
     \label{eqn:S1S2Avg}
\end{equation}
where {$n_{3}$ is the number of classes $N_{C}$ and $n_4 = {N_C \choose 2} = N_C(N_C - 1)/2$} is the number of combinations of class pairs without repetition.



{Our earlier results (summarized in Table \ref{tab:group_sim} in later sections) show that classifying a dataset is difficult when images in a class are similar to other classes. This observation implies that a dataset's difficulty is directly related to inter-class similarity and inversely to the intra-class similarity. Based on this, our difficulty measure design rationale is to capture both types of similarity jointly, summarized as \textbf{weighted similarity score $\bar{S}$}. To further ensure the score falls in a consistent range, we additionally design a measure, dubbed \textbf{soft similarity score $\hat{S}$} that normalizes the weighted intra- and inter-class similarity scores by its maximum. Formally,}
 we define {$\bar{S}$ and $\hat{S}$ as,}
{
\begin{equation}
    \hspace{-5px}\bar{S} = \frac{1 + \lambda_{s}\bar{S}_{R} - (1-\lambda_{s})\bar{S}_{E}}{2}, \hspace{27px} \hat{S} = \frac{\lambda_{s}\bar{S}_{R} - (1-\lambda_{s})\bar{S}_{E}}{max(\lambda_{s}\bar{S}_{R}, (1-\lambda_{s})\bar{S}_{E})}
    \label{equ:s}
\end{equation}
}
{where $\lambda_{s}$ is a weighting factor (default to $0.5$) to balance $\bar{S}_{R}$ and $\bar{S}_{E}$. }





\section{Experiment}
\label{sec:Exp}

In this section, we begin with experiments showing the effect of the number of classes on model accuracy. We then add similarity, and in the last subsection show how the combined difficulty measure is used for a real application.

\subsection{Number of Classes}
\label{sec:numClasses}

It is known that accuracy reduces when more classes are involved, or equivalently a larger model is needed to maintain the same accuracy. This is because more visual features are needed to separate the classes and the decision boundary is more complex accordingly. We perform experiments in this section, first to confirm this relationship empirically, and second to gain a more quantitative insight into how the relationship changes across the range of few to more classes.

\begin{figure*}[t]
  \begin{minipage}{1\linewidth}
  \centering
    \subfigure[$N_{CL}$ on CIFAR10]{\includegraphics[width=0.325\textwidth]{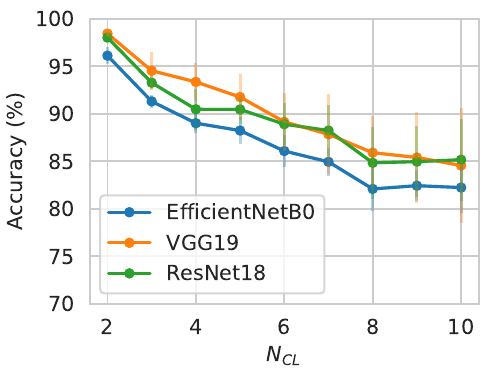}}
    \subfigure[$N_{CL}$ on COCO]{\includegraphics[width=0.325\textwidth]{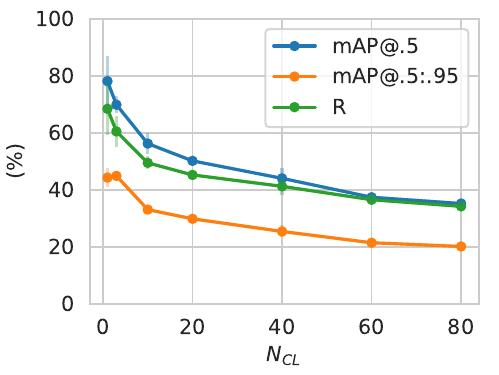}}
    \subfigure[sub-YOLO on COCO]{\includegraphics[width=0.325\textwidth]{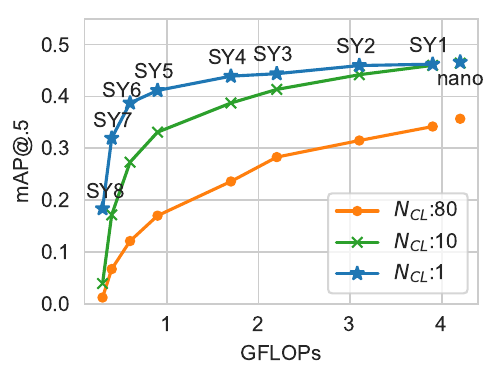}}
  \caption{Overall relationship of performance versus the number of classes $N_{CL}$. {Each dot denotes an average of 3 subsets for each $N_{CL}$, while error bars represent standard deviations (each multiplied by 5 in (b) for visibility).} (a) Image classification accuracy decreases for the classifiers tested when the number of CIFAR-10 classes is increased from 2 to 10. (b) Object detection accuracy and recall (R) decrease when the number of COCO classes is increased from 1 to 80.  (c) Accuracy plot for increasingly smaller models from YOLOv5-nano through eight sub-YOLO models (SY1-8) and  class groupings of 1 ($N_{CL}$:1), 10 ($N_{CL}$:10), and 80 ($N_{CL}$:80).}
  \label{fig:n_cls_plots}
  \end{minipage}
\end{figure*}

We performed {four} sets of experiments. The first was for object detection using the YOLOv5-nano ~\cite{glenn_jocher_2021_5563715} backbone upon randomly-chosen, increasing-size class groupings of the COCO dataset ~\cite{lin2014microsoft}. {Ten groups with $N_{CL}$ of $\{$1, 2, 3, 4, 5, 10, 20, 40, 60, 80$\}$ were prepared. For each group, we trained a separate YOLOv5-nano model from scratch. we set the initial learning rate as $0.01$ with weight decay $0.0005$ at image size $640$ using SGD optimizer.} As seen in Fig. \ref{fig:n_cls_plots} (a), accuracy decreases with number of classes as expected. But perhaps not anticipated is the fact that the accuracy decrease is steep for very few classes, say 5-10 or fewer, and flattens beyond 10. We will examine later in the paper the difference between classification with few versus more classes.

The second set of experiments is for image classification on the CIFAR-10 dataset. With many fewer classes in CIFAR-10 ~\cite{krizhevsky2009learning} than COCO (10 versus 80), we expect to see how the number of classes and accuracy relate for this smaller range. We extracted subsets of classes --- which we call groups -- from CIFAR-10 with $N_{CL}$ ranging from 2 to 9. For example, one group with $N_{CL}=4$ contains \textit{airplane}, \textit{cat}, \textit{automobile}, and \textit{ship} classes. We trained 3 classifiers from scratch for each group, EfficientNet-B0, VGG19 \cite{vggNet2014}, and MobileNet V2 \cite{sandler2018mobilenetv2}. Results of the image classification experiments are shown in Fig. \ref{fig:n_cls_plots} (middle). The classifiers used for testing, showed the expected trend of accuracy reduction as $N_{CL}$ per group increased. However, the trend was not as monotonic as might be expected. We hypothesized that this might be due to the composition of each group. Class groupings were randomly chosen, so they have different levels of inter-class similarity. We explore how inter-class similarity affects accuracy in the next section.

The third set of experiments involves reducing model size for classifying different numbers of classes and measuring accuracy versus computation effort in GFLOPS. We prepared 90 random class groups from the COCO minitrain dataset ~\cite{HoughNet}. There are 80 groups with $N_{CL}=1$, each containing a single class from 80 classes. There are 8 groups with $N_{CL}=10$. The final dataset is the original COCO minitrain with $N_{CL}=80$.
We scale YOLOv5 layers and channels down in model size with the depth and width factors already used for scaling the family up in size from nano to x-large. Starting with depth and width multiples of 0.33 and 0.25 for YOLOv5-nano, we reduce these in step sizes of 0.04 for depth and 0.03 for width. In this way, we design a monotonically decreasing sequence of sub-YOLO models denoted as SY1 to SY8. We train each model separately for each of the six groupings. Results of sub-YOLO detection are shown in Fig. \ref{fig:n_cls_plots} (right). There are three lines where each point of $mAP@.5$ is averaged across all models in all datasets for a specific $N_{CL}$. An overall trend is observed that fewer-class models (upper-left blue star) achieve higher efficiency than many-class models. Another important finding is that, whereas the accuracies for 80 classes drops steadily from the YOLOv5-nano size, accuracy for 10 classes is fairly flat down to SY2, which corresponds to a 36\% computation reduction, and for 1 class down to SY4, which corresponds to a 72\% computation reduction.

\begin{figure*}[h]
  \begin{minipage}{1\linewidth}
  \centering
    {\includegraphics[width=0.325\textwidth]{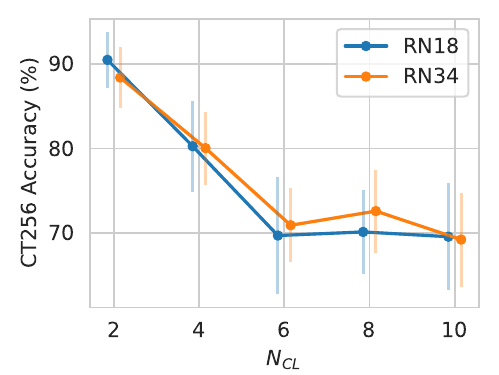}}
    \vspace{-9pt}
    {\includegraphics[width=0.325\textwidth]{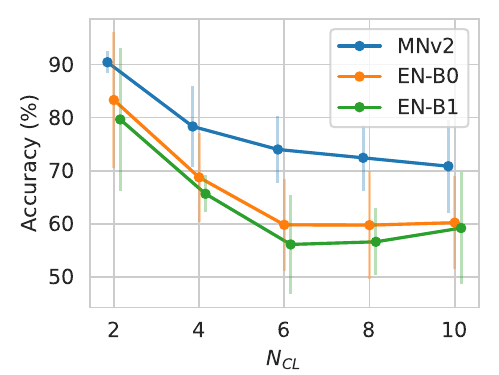}}
    {\includegraphics[width=0.325\textwidth]{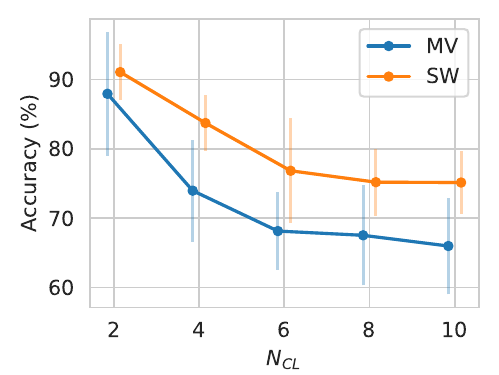}}
    \vspace{-9pt}
    {\includegraphics[width=0.325\textwidth]{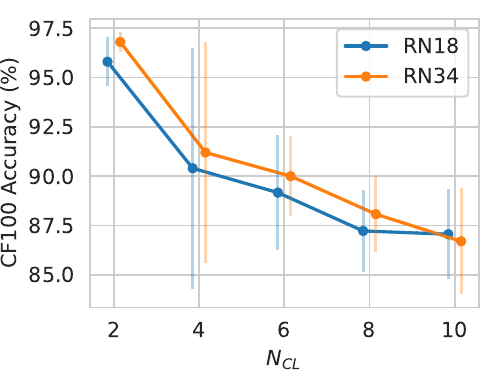}}
    {\includegraphics[width=0.325\textwidth]{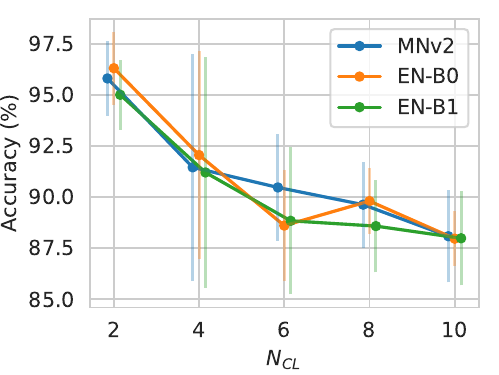}}
    {\includegraphics[width=0.325\textwidth]{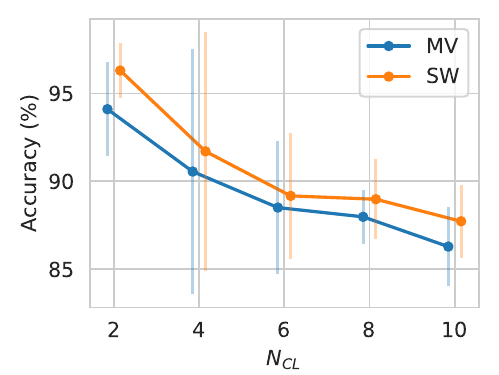}}
    \vspace{-9pt}
    {\includegraphics[width=0.325\textwidth]{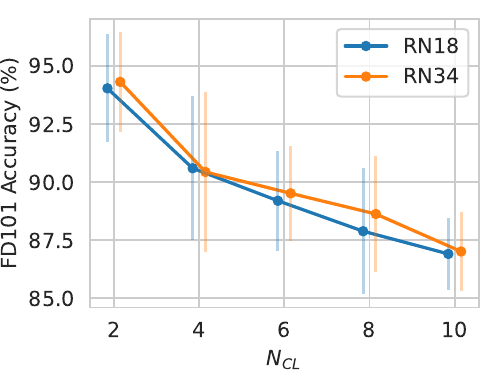}}
    {\includegraphics[width=0.325\textwidth]{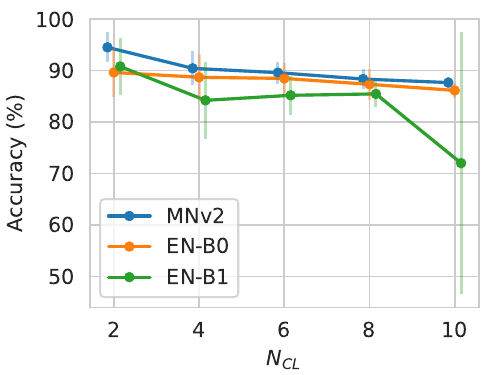}}
    {\includegraphics[width=0.325\textwidth]{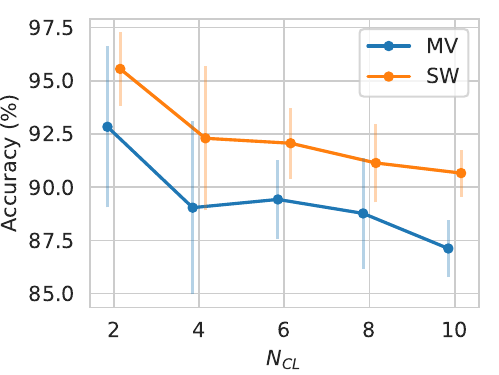}}
    {\includegraphics[width=0.325\textwidth]{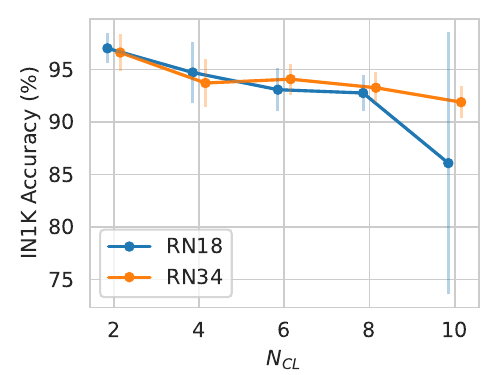}}
    {\includegraphics[width=0.325\textwidth]{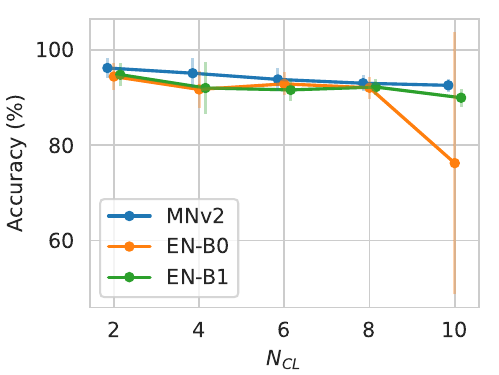}}
    {\includegraphics[width=0.325\textwidth]{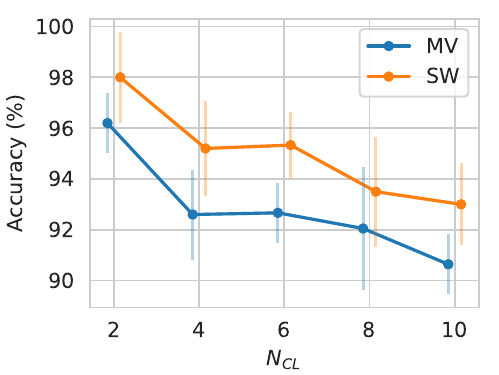}}
  \caption{{Overall trend of 5 efficient CNNs and 2 ViTs on 4 datasets: model performance tends to decrease while $N_{CL}$ increases. Each dot denotes the average accuracy of 5 subsets for $N_{CL}$. The error bars represent standard deviations of accuracy in 5 subsets. Accuracy: Top-1 Accuracy. RN: ResNet. MNv2: Mobilenet V2. EN: EfficientNet. MV: MobileViT. SW: Swin Transformer. CT256: CalTech256, CF100: CIFAR100, FD101: Food101, IN1K: ImageNet1K.}}
  \label{fig:4ds}
  \end{minipage}
\end{figure*}

{In the fourth set of experiments, we extend to 7 models on 4 datasets, covering 5 Convolutional Neural Networks (CNNs) and 2 Vision Transformers (ViTs) shown in Fig. \ref{fig:4ds}, totaling 1225 training/testing runs. Models include ResNet18, ResNet34 \cite{he2016deep}, MobileNet V2 \cite{sandler2018mobilenetv2}, EfficientNet-B0, EfficientNet-B1 \cite{tan2019efficientnet}, MobileViT \cite{mehta2021mobilevit} and Swin Transformer \cite{liu2021swin}. Transformers are included for completeness. However, they are initially designed for scaling up, which is not efficient as the focus of this paper. Therefore a lightweight ViT MobileViT is chosen. We select datasets consisting of natural images, including CalTech256 \cite{griffin_holub_perona_2022}, CIFAR100 \cite{krizhevsky2009learning}, Food101 \cite{bossard14} and ImageNet1K \cite{deng2009imagenet}. We vary the number of classes $N_{CL}$ from $2$ to $10$ with step size $2$. For each $N_{CL}$, 5 subsets are randomly selected (seed number 0-4) from the full dataset. A model is trained and converged in each subset, then the validation Top-1 accuracy is reported. By default, we use SGD optimizer with learning rate $0.1$, momentum $0.9$ and weight decay $0.0001$ to train our models.
 }


The importance of the findings in this section should be emphasized for few-class, practical applications where computational efficiency and low energy use is required. Not only is accuracy higher for fewer-class application datasets, but the practitioner can choose smaller and smaller models with small accuracy penalty, but much smaller energy use (as shown in the right plot of Fig. \ref{fig:n_cls_plots}).



\subsection{Intra- and Inter-Class Similarity}
\label{sec:Exp_intersim}

We start with an experiment for didactic purposes to indicate how our subjective notion of similarity relates to accuracy. We test with 3 models, EfficientNet-B0 (EB0), VGG-19 (V19), and MobileNet V2 (MV2). Table \ref{tab:group_sim} shows accuracy results for groupings of 2 and 4 classes from the CIFAR-10 dataset, which we have subjectively attributed similarity values of ``yes'' and ``no''. For each model, the accuracies in bold are the highest for their groupings, corresponding to the class groupings of lower subjective similarity. In the final column (I-CS), the objective inter-class similarity scores also correspond to our subjective designations.

\begin{table}[h]
  \centering
  \begin{tabular}{@{}c c c c c c c}
   \toprule
    nCl & Classes & Similarity & EB0 & V19 & MV2 & I-CS \\
   \specialrule{.10em}{.05em}{.05em}
    4 & cat, deer, dog, horse & yes  & 0.84 & 0.86 & 0.76 & 0.57 \\
    \midrule
    4 & airplane, cat, auto, ship & no &  \textbf{0.91} & \textbf{0.94} & \textbf{0.93} & 0.12 \\
   \specialrule{.10em}{.05em}{.05em} 
    2 & deer, horse & yes & 0.92 & 0.94 & 0.89 & 0.61 \\
    2 & auto, truck &  yes & 0.91 & 0.95 & 0.93 & 0.56 \\
    \midrule
    2 & airplane, frog & no &  \textbf{0.98} & \textbf{0.98} & \textbf{0.96} & 0.11 \\
    2 & deer, ship & no & \textbf{0.98} & \textbf{0.98} & \textbf{0.96} & 0.08 \\
   \bottomrule
  \end{tabular}
  \caption{{Accuracies for three image classifiers (EB0, V19, MV2) for class groupings of $nCL = 2 \textrm{ and } 4$ whose similarities are initially subjectively assigned, but are supported by Inter-Class Similarity (I-CS), an objective measurement in the final column.}}
  \label{tab:group_sim}
\end{table}

\begin{figure*}[t]
  \centering
   \subfigure[{Inter-Class \textbf{Sim.}}]{\includegraphics[width=0.346\linewidth]{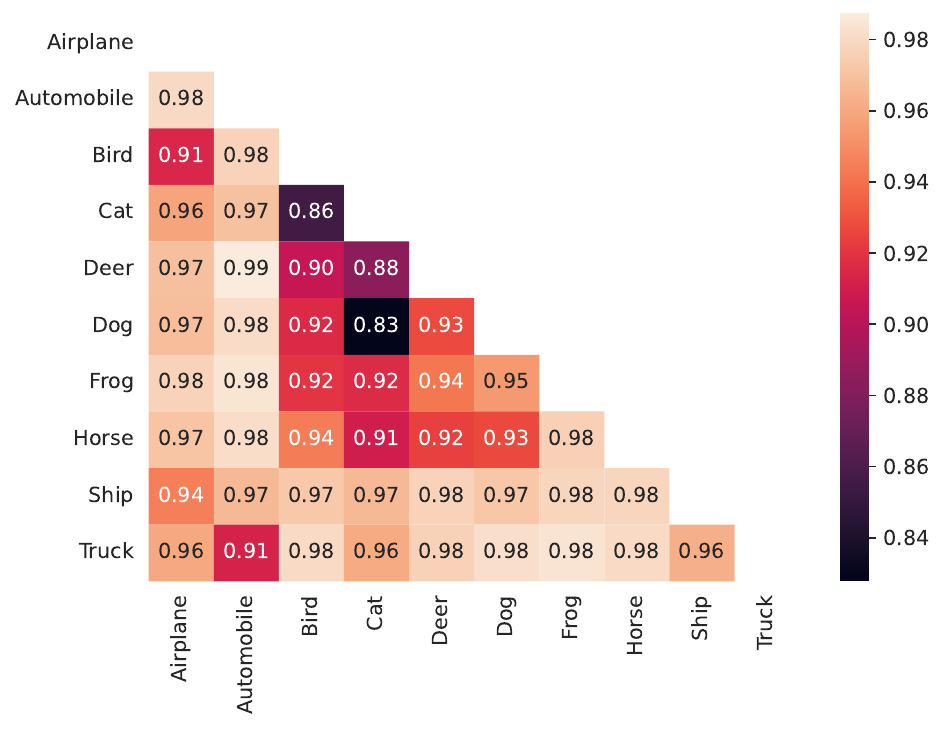}}
    \subfigure[{Bi-Classification \textbf{Acc.}}]{\includegraphics[width=0.346\linewidth]{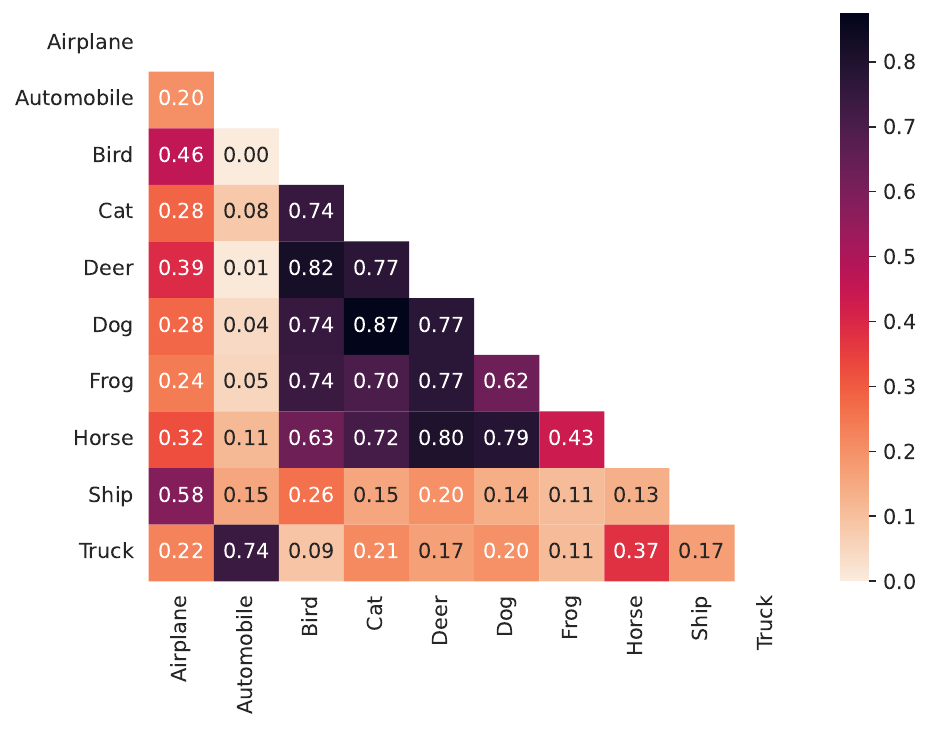}}
    \subfigure[{Rel. of \textbf{Sim.} \& \textbf{Acc.}}]{\includegraphics[width=0.288\linewidth]{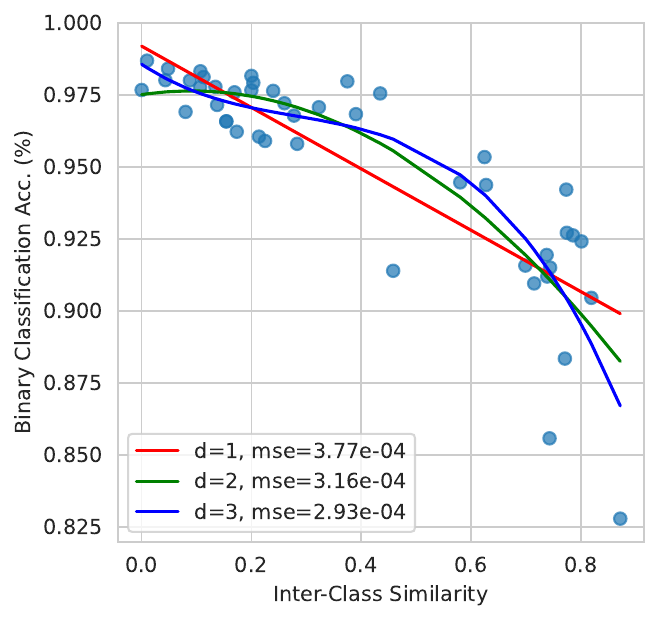}}
  \caption{Matrices showing relationships among pairs of classes: (a) binary classification accuracy matrix using EfficientNet-B0, (b) binary-class similarity matrix with $S_E$ metric, (c) {nonlinear} relationship between binary classification accuracy (a) and similarity scores (b). {The polynomial function with a degree of $d=3$ (blue) has least mse compared with $d=2$ (green) and $d=1$ (red). Sim.: Similarity. Acc.: Accuracy. Rel.: Relationship.}}
  \label{fig:sim_matrix}
\end{figure*}

To illustrate the relationship between class similarity scores and accuracy, we calculate these using EfficientNet-B0 for all pairwise classifications in CIFAR-10 and show results in Fig. \ref{fig:sim_matrix}. The left matrix shows accuracy results between pairs of classes. The middle matrix shows average intra-class similarity scores on the diagonal and inter-class scores in the off-diagonal boxes. The absolute value of the Pearson correlation coefficient \cite{benesty2009pearson} {|r|} between the binary classification accuracy (left matrix) and the similarity scores (middle matrix) is \textbf{0.77}, showing \textbf{strong correlation} between these two measures. All (similarity, accuracy) pairs are shown in the plot (right) clearly indicating the strong inverse correlation. Of note, the lowest similarity data point in the top left of the plot is for the (automobile, deer) pair, and the highest similarity data point in the bottom right is for the (cat, dog) pair.

{We further explore the type of relationship between similarity and accuracy in terms of nonlinearity. Specifically, in the data points in Fig. \ref{fig:sim_matrix} (c), we fit three polynomial functions $f(s) = p_{0}x^{d} + ... + p_{(d-1)}x + p_{d}$ with degree $d \in \{1, 2, 3\}$. To measure these functions accuracy, we compute the Mean Squared Error $mse=\frac{1}{N}\sum_{i}^{N}(acc^{\prime}_{i} - acc_{i})^2 $ between the prediction of accuracy ($acc^{\prime}$) and ground truth of accuracy ($acc$) with inter-class similarity as input. Results show that degree of $3$ has the least $mse=2.93\times10^{-4}$ compared to degrees of $2$ $ (mse=3.16\times10^{-4})$ and $1$ $(mse=3.77\times10^{-4})$, respectively. The nonlinearity (e.g. $d=3$) between similarity and accuracy can suggest that when images in a dataset become similar and similar in a certain scale, the expected gain of a model's accuracy is diminished. This insight can assist the estimation of an efficient neural network's performance in a real-world scenario under some resource-constraints.}


\subsection{Similarity Metric Efficiency}
\label{sec:similarityEfficiency}

To select a neural network model for an application, a common first step is to identify models that attain the requirements. In this paper, we focus on efficient applications, so appropriate selections include small model families such as YOLO, MobileNet, and EfficientNet-B0 (classification or object detection variants of these depending upon the goal of the application). A model would be chosen, trained and tested. Another model, larger or smaller depending upon results of the first choice, would be trained and tested, and compared. This training and testing cycle would continue until the model closest the the application requirements is found.

Use of the difficulty measure offers a much faster way to find an appropriate model. If a model of interest has been previously trained and tested on an application, and the difficulty measure of that application has been determined, then the practitioner can do the following. Calculate the difficulty measure of the current application dataset. If the measure is higher than previously, then the application will likely require a larger model, and if not a smaller model. This procedure guides the selection process up or down the model family levels, where the advantage is that training and testing are not required, just similarity score measurement. 

Note that this is a relative measurement procedure, which needs to start with a model already trained-and-tested, but for a developer who has done many applications, this is already available. Furthermore, one could project that practitioners would publish similarity scores and associated model accuracies for applications they have done for the benefit of other practitioners. To be clear, the application classes do not have to be the same, just the relative difficulty measures.

The traditional alternative is to perform training and inference testing all combinations of attribute values. For $N_{CL}$ classes, binary classification of pairs requires $N_{CL} \choose 2$ training and inference operations. In comparison, for the difficulty measure, we need to train once for all combinations of binary classifications (the same training that would happen traditionally). Then, instead of testing, only the pairwise similarities between pairs needs by performed, an expense of one vector multiply each, rather than a full neural network test requiring the number of multiplies of the model. For subsequent tests of different datasets, training does not have to be repeated; instead the cached latent space is used for pairwise similarity tests.

An example of runtime comparison for the CIFAR10 dataset is shown in Table ~\ref{tab:sim_eff}. We compare against three convolutional neural networks (CNNs) designed for small or embedded applications. Both CNN and similarity metric approaches require initial training. For the similarity metric, there is an additional one-time task of feeding all test images into the similarity metric, which takes 0.72 seconds, and then caching, which takes 0.63 seconds. Applying the similarity metric takes 0.76 seconds to calculate a similarity score for each pair of images. In comparison for conventional testing, each image must undergo full CNN testing to obtain its predicted accuracy.

\begin{table}[h]
  \centering
  \label{tab:sim_eff}
\begin{tabular}{p{100pt} p{80pt} p{50pt}}
    \toprule
    Model & $t_{train}$ (s/epoch) & $t_{test}$ (s/pair) \\
    \midrule
    VGG19 & 0.69 & \hspace{1pt} 4.49 \\
    EfficientNet-B0 & 3.13 & 21.82 \\
    MobileNet V2 & 2.19 & 15.05 \\
    \textbf{Similarity Metric} & 3.31 & \hspace{-1pt} \textbf{0.76} \\
    \bottomrule
  \end{tabular}
  \caption{{Results of runtime comparison of Similarity Metrics and some popular CNN models. s: second, pair: all pair of instances in two classes.}}
\end{table}

\subsection{{Difficulty Measure Evaluation}}

{The purpose of a difficulty measure is to provide a means to estimate a model's performance. To that end, we employ the Pearson correlation coefficient \cite{benesty2009pearson} $r$ to measure the correlation between difficulty and accuracy. A higher value of the absolute value $|r|$ indicates a stronger correlation of two vectors, while \textbf{$|r| > 7$} is commonly considered as \textbf{strongly correlated} \cite{pcc} \cite{schober2018correlation}. Since we target efficiency, we use the smallest models in ResNet18 and EfficientNet-B0 in this experiments. CalTech256 is selected due to its focused object categories, balanced class distribution and consistent labels, which are suitable for resource-constrained research.}

{In the following experiments, similarity is computed in the latent space of image encoder in DINOv2 \cite{oquab2023dinov2} -- a large-scale model that is commonly used to extract robust image features in recent research.}


{\textbf{Difficulty Method Comparison:} We compare our proposed similarity-based measures $\bar{S}$ and $\hat{S}$ against Euclidean distance baselines  $\bar{D}$ and $\hat{D}$ in DINOv2 latent space, where,}
{
\begin{equation}
    \bar{D} = \frac{1 + \lambda_{d}\bar{D}_{E} - (1-\lambda_{d})\bar{D}_{R}}{2}, \hspace{20px} \hat{D} = \frac{\lambda_{d}\bar{D}_{E} - (1-\lambda_{d})\bar{D}_{R}}{max(\lambda_{d}\bar{D}_{E}, (1-\lambda_{d})\bar{D}_{R})}
\end{equation}
}

{Results in Table \ref{tab:pcc_main} demonstrate that our proposal $S$ ($|r|>0.7$) outperforms the baseline $D$ ($|r|<0.7$).}

\begin{table}[h]
  \centering
\begin{tabular}{p{30pt}|p{30pt}p{65pt}|p{30pt}p{65pt}}
    \toprule
    & \multicolumn{4}{c}{|r|} \\
    \cmidrule(r){2-5}
    $\lambda$ & $\bar{D}$ & $\bar{S}$ & $\hat{D}$ & $\hat{S}$ \\
    \midrule
    0.25 & 0.546 & \textcolor{red}{0.757 (+0.211)} & 0.600 & \textcolor{red}{0.788 (+0.188)} \\
    0.50 & 0.696 & \textcolor{red}{0.789 (+0.093)} & 0.660 & \textcolor{red}{0.796 (+0.136)} \\
    0.75 & 0.691 & \textcolor{red}{0.762 (+0.071)} & 0.660 & \textcolor{red}{0.796 (+0.136)} \\
    \bottomrule
  \end{tabular}
  \caption{{Comparison of similarity-based $S$ and Euclidean distance-based $D$ difficulty measure with various $\lambda$ by the absolute value of Pearson correlation coefficient $|r|$.}}
  \label{tab:pcc_main}
\end{table}

{\textbf{Effect of $\lambda$:} We vary the weight $\lambda$ that balances $S_{R}$ and $S_{E}$ ($D_{R}$ and $D_{E}$ for $D$. Results in Table \ref{tab:pcc_main}) show that both reach the highest scores when $\lambda=0.5$, specifically $\bar{S}=0.789$, $\hat{S}=0.796$, $\bar{D}=0.696$, $\hat{D}=0.660$, respectively.}

{\textbf{Difficulty Measure Ablation Study:} We ablate components in $\bar{S}$ defined in Equation \ref{equ:s}. In particular, ablating $\bar{S}_{R}$ reduces the absolute Pearson correlation coefficient $|r|$ from $0.789$ to $0.692$, while $|r|$ decreases to $0.702$ by removing $\bar{S}_{E}$.}

\begin{table}[h]
  \centering
\begin{tabular}{p{90pt}p{90pt}|p{65pt}}
    \toprule
    $w/o \hspace{3pt} \bar{S}_{R} \hspace{3pt} (\lambda_{s}=0) $ & $w/o \hspace{3pt} \bar{S}_{E} \hspace{3pt} (\lambda_{s}=1)$ & $\bar{S}$ \\
    \midrule
    0.692 & 0.702 & \textcolor{red}{0.789} \\
    \bottomrule
  \end{tabular}
  \caption{{Ablation Study of $\bar{S}$ by Pearson correlation coefficient |r|.}}
  \label{tab:abl}
\end{table}

{Results in both Table \ref{tab:pcc_main} and \ref{tab:abl} verify our design intuition that jointly considering intra- and inter-class difficulty measures is beneficial.}

Besides helping select an appropriate model there is another practical use for difficulty measurement of an application dataset. Sometimes the requirements for an application offer options to enable trading off one requirement for another. For example, application requirements might allow two options, one for classification of 5 classes at some accuracy level and cost ceiling, and another for classification of 4 classes at the same accuracy level but at a lower cost. An industry example of this requirement tradeoff is given in Section \ref{sec:Appl_robot}.

\section{Classication Difficulty for an Industry Application}
\label{sec:Appl_robot}
We have applied the classification difficulty measure to model selection for an industry application involving video analytics for human-robot interaction. 

\begin{table*}[h]
  \centering
  \footnotesize
  \begin{tabular}{c c c c c c c c c}
    \toprule
    $N_{CL}$ & S & Class & Ym & Ys & Yn & Y-1 & Y-2 & Y-3\\
     & & Label & (21.1M) & \textcolor{red}{(7.2M)} & (1.9M) & \textcolor{red}{(1.1M)} & (0.16M) & (0.07M) \\
    \midrule
    &  & p-walk & 0.748 & 0.728 & 0.727 & 0.725 & 0.665 & 0.654 \\
    3 & 0.18 & p-cart & 0.720 & 0.690 & 0.674 & 0.681 & 0.576 & 0.500 \\
    & & robot & 0.865 & 0.872 & 0.827 & 0.82 & 0.747 & 0.635 \\
    \cmidrule{3-9}
    & & mAP & 0.778 & \textcolor{red}{0.764} & 0.743 & 0.742 & 0.663 & 0.596 \\
    \cmidrule{1-9}
    &  & person & 0.753 & 0.752 & 0.732 & 0.719 & 0.691 & 0.657 \\
    2 & 0.15 & robot & 0.872 & 0.875 & 0.827 & 0.814 & 0.755 & 0.650 \\
     \cmidrule{3-9}
     & & mAP & \textbf{0.812} & \textbf{0.813} & \textbf{0.779} & \textcolor{red}{\textbf{0.766}} & \textbf{0.723} & \textbf{0.654} \\
     & & & 4.4\% & 6.4\% & 4.9\% & 3.2\% & 9.1\% & 9.7\% \\
    \bottomrule
  \end{tabular}
  \caption{Results of detection of YOLOv5 medium (Ym), small (Ys), nano (Yn) and the sub-YOLO models (Y-1, Y-2, Y-3) on 3-class (person-walk, person-cart, robot) and 2-class (person, robot) cases respectively. The bracketed numbers below the model names are their model sizes. The accuracy is mAP@0.5. The numbers in the second from bottom row are in bold to show that 2-class accuracy is higher than for 3-class, and the numbers in the bottom row show the percentage improvement. The red numbers show that the sub-YOLO1 model can achieve similar accuracy to the YOLOs model, but with $6.5\times$ smaller size when class grouping is reduced from 3 to 2. The bottom row shows the accuracy improvement from 3 to 2 classes for each model.}
  \label{tab:YOLO_table}
\end{table*}

The objective is to recognize human activity from fixed hallway cameras of an assembly factory so as to reduce human-robot interaction (HRI). Three classes were identified and trained for this application, \textit{person-walk}, \textit{person-cart}, and \textit{robot}. The \textit{person} class was initially separated into two, \textit{person-walk} and \textit{person-cart} (person walking and person pushing a cart). This was because this distinction was deemed useful -- and we didn't want to retrain if we just trained on two classes at the outset.

Results in Table \ref{tab:YOLO_table} show in general that the 3-class group with similarity value 0.18 has lower accuracy across models than the 2-class group with similarity value 0.15. For the 3-class option, one good choice that balances accuracy and size would be the sub-YOLO1 model, whose accuracy is just $0.743 - 0.742 = 0.001$ less than the YOLO-nano model, but whose size is $1.1 / 1.9 = 0.579$ (or $42\%$) of the YOLO-nano. When the \textit{person-walk} and \textit{person-cart} classes are merged into a single \textit{person} class, then sub-YOLO1 could be chosen with essentially the same accuracy as YOLOs, but with 85\% smaller size.

\section{Conclusion}
\label{sec:Conclusion}

The difficulty measure proposed here provides a relative measure that, knowing the performance of a model for one dataset, one can predict the model performance for the same dataset on different models of a model family or on other datasets on the same model.


In this paper, we have proposed a measure of dataset classification difficulty based upon three characteristics of a dataset, number of classes, intra-class similarity, and inter-class similarity. We have experimented with 9 neural network models on 7 datasets to demonstrate the relationship between model accuracy and dataset difficulty. Our proposed similarity-based method outperforms the baseline using Euclidean distance in terms of correlation with accuracy by Pearson correlation coefficient. We have shown the utility of the difficulty measure in guiding a practitioner to an efficient model architecture without repeated training and testing for different datasets.

\section{Acknowledgement}
\label{sec:ack}
This research has been supported in part by the National Science Foundation (NSF) under Grant No. CNS-2055520.

\bibliography{main}

\begin{thebibliography}{10}
\providecommand{\url}[1]{\texttt{#1}}
\providecommand{\urlprefix}{URL }
\providecommand{\doi}[1]{https://doi.org/#1}

\bibitem{appalaraju2017image}
Appalaraju, S., Chaoji, V.: Image similarity using deep cnn and curriculum learning. arXiv preprint arXiv:1709.08761  (2017)

\bibitem{bellet2013survey}
Bellet, A., Habrard, A., Sebban, M.: A survey on metric learning for feature vectors and structured data. arXiv preprint arXiv:1306.6709  (2013)

\bibitem{benesty2009pearson}
Benesty, J., Chen, J., Huang, Y., Cohen, I.: Pearson correlation coefficient. In: Noise reduction in speech processing, pp.~1--4. Springer (2009)

\bibitem{bossard14}
Bossard, L., Guillaumin, M., Van~Gool, L.: Food-101 -- mining discriminative components with random forests. In: European Conference on Computer Vision (2014)

\bibitem{applDriving2021}
Cai, Y., Luan, T., Gao, H., Wang, H., Chen, L., Li, Y., Sotelo, M.A., Li, Z.: Yolov4-5d: An effective and efficient object detector for autonomous driving. IEEE Trans. on Instrumentation and Measurement  \textbf{70},  1--13 (2021)

\bibitem{ScaledInference_2022}
Chang, Y.J., Hong, D.Y., Liu, P., Wu, J.J.: Efficient inference on convolutional neural networks by image difficulty prediction. In: 2022 IEEE Int. Conf. on Big Data (Big Data). pp. 5672--5681 (2022)

\bibitem{deng2009imagenet}
Deng, J., Dong, W., Socher, R., Li, L.J., Li, K., Fei-Fei, L.: Imagenet: A large-scale hierarchical image database. In: 2009 IEEE conference on computer vision and pattern recognition. pp. 248--255. Ieee (2009)

\bibitem{pedestrian2017}
Du, X., El-Khamy, M., Lee, J., Davis, L.: Fused \uppercase{DNN}: A deep neural network fusion approach to fast and robust pedestrian detection. In: 2017 IEEE Winter Conf. on Applications of Computer Vision (WACV). pp. 953--961 (2017)

\bibitem{SimilarityMetricLearning2021}
Duffner, S., Garcia, C., Idrissi, K., Baskurt, A.: Similarity Metric Learning, pp. 103--125. Springer Int. Publishing, Cham (2021)

\bibitem{hfhub}
Face, H.: Hugging face model hub (August 22 2023), \url{https://huggingface.co/models}, accessed on 2023-08-22

\bibitem{pytorchhub}
Foundation, T.L.: Pytorch model hub (August 22 2023), \url{https://pytorch.org/hub/}, accessed on 2023-08-22

\bibitem{ganesh2022yolo}
Ganesh, P., Chen, Y., Yang, Y., Chen, D., Winslett, M.: Yolo-ret: Towards high accuracy real-time object detection on edge gpus. In: Proc. IEEE/CVF Winter Conf. on Applications of Computer Vision. pp. 3267--3277 (2022)

\bibitem{griffin_holub_perona_2022}
Griffin, G., Holub, A., Perona, P.: Caltech 256 (Apr 2022). \doi{10.22002/D1.20087}

\bibitem{SommelierModelRepository_2022}
Guo, P., Hu, B., Hu, W.: Sommelier: Curating \uppercase{DNN} models for the masses. In: Proc. 2022 Int. Conf. on Management of Data. p. 1876–1890. SIGMOD '22, Association for Computing Machinery (2022)

\bibitem{medDiagnose_2022}
Hannemose, M.R., Sundgaard, J.V., et~al.: Was that so hard? estimating human classification difficulty. In: Wu, S., et~al. (eds.) Appl.s of Medical Artificial Intelligence. pp. 88--97. Springer Nature Switzerland (2022)

\bibitem{he2016deep}
He, K., Zhang, X., Ren, S., Sun, J.: Deep residual learning for image recognition. In: Proceedings of the IEEE conference on computer vision and pattern recognition. pp. 770--778 (2016)

\bibitem{ClassComplexityHo2002}
Ho, T.K., Basu, M.: Complexity measures of supervised classification problems. IEEE Trans. on Pattern Analysis and Machine Intelligence  \textbf{24}(3),  289--300 (2002)

\bibitem{howard2017mobilenets}
Howard, A.G., Zhu, M., Chen, B., Kalenichenko, D., Wang, W., Weyand, T., Andreetto, M., Adam, H.: Mobilenets: Efficient convolutional neural networks for mobile vision applications. arXiv preprint arXiv:1704.04861  (2017)

\bibitem{tfhub}
Inc., G.: Tensorflow model hub (August 22 2023), \url{https://www.tensorflow.org/hub}, accessed on 2023-08-22

\bibitem{appl_cattle}
J.G.A.~Barbedo, L.V.~Koenigkan, T.S.P.S.: A study on the detection of cattle in uav images using deep learning. Sensors  \textbf{19}(24) (2019)

\bibitem{ZestForLIME_2022}
Jia, H., Chen, H., Guan, J., Papernot, N.: A zest for \uppercase{LIME}: Toward architecture-independent model distances. In: {ICLR 2022 - 10th Int. Conf. on Learning Representations}. p. 1876–1890. Virtual, France (Apr 2022)

\bibitem{glenn_jocher_2021_5563715}
Jocher, G., et. al.: {ultralytics/yolov5: v6.0 - YOLOv5n 'Nano' models, Roboflow integration, TensorFlow export, OpenCV \uppercase{DNN} support} (Oct 2021)

\bibitem{koonce2021mobilenetv3}
Koonce, B., Koonce, B.: Mobilenetv3. Convolutional Neural Networks with Swift for Tensorflow: Image Recognition and Dataset Categorization pp. 125--144 (2021)

\bibitem{koonce2021squeezenet}
Koonce, B., Koonce, B.: SqueezeNet. Springer (2021)

\bibitem{krizhevsky2009learning}
Krizhevsky, A., Hinton, G., et~al.: Learning multiple layers of features from tiny images  (2009)

\bibitem{lin2014microsoft}
Lin, T.Y., Maire, M., Belongie, S., Hays, J., Perona, P., Ramanan, D., Doll{\'a}r, P., Zitnick, C.L.: Microsoft coco: Common objects in context. In: European Conf. on computer vision. pp. 740--755. Springer (2014)

\bibitem{appl_shipSAR}
Liu, S., Kong, W., Chen, X., Xu, M., Yasir, M., Zhao, L., Li, J.: Multi-scale ship detection algorithm based on a lightweight neural network for spaceborne sar images. Remote Sensing  \textbf{14}(5) (2022)

\bibitem{liu2021swin}
Liu, Z., Lin, Y., Cao, Y., Hu, H., Wei, Y., Zhang, Z., Lin, S., Guo, B.: Swin transformer: Hierarchical vision transformer using shifted windows. In: Proceedings of the IEEE/CVF international conference on computer vision. pp. 10012--10022 (2021)

\bibitem{metricLearn2017}
Lu, J., Hu, J., Zhou, J.: Deep metric learning for visual understanding: An overview of recent advances. IEEE Signal Processing Magazine  \textbf{34}(6),  76--84 (2017)

\bibitem{appl_crowd}
M.~Sabokrou, M.~Fayyaz, M.F.Z.M.R.K.: Deep-anomaly: Fully convolutional neural network for fast anomaly detection in crowded scenes. Computer Vision and Image Understanding  \textbf{172},  88--97 (2018)

\bibitem{ma2018shufflenet}
Ma, N., Zhang, X., Zheng, H.T., Sun, J.: Shufflenet v2: Practical guidelines for efficient cnn architecture design. In: Proc. European Conf. on computer vision (ECCV). pp. 116--131 (2018)

\bibitem{weightedModel2020}
Marsden, M., McGuinness, K., et~al.: Investigating class-level difficulty factors in multi-label classification problems. In: 2020 IEEE Int. Conf. on Multimedia and Expo (ICME). pp.~1--6 (2020)

\bibitem{imagenetDifficulty2022}
Meding, K., Buschoff, L.M.S., Geirhos, R., Wichmann, F.A.: Trivial or impossible --- dichotomous data difficulty masks model differences (on imagenet and beyond). In: Int. Conf. on Learning Representations (2022)

\bibitem{mehta2021mobilevit}
Mehta, S., Rastegari, M.: Mobilevit: light-weight, general-purpose, and mobile-friendly vision transformer. arXiv preprint arXiv:2110.02178  (2021)

\bibitem{applWildlifeAustralia2017}
Nguyen, H., Maclagan, S.J., Nguyen, et~al.: Animal recognition and identification with deep convolutional neural networks for automated wildlife monitoring. In: 2017 IEEE Int. Conf. on Data Science and Advanced Analytics. pp. 40--49 (2017)

\bibitem{oquab2023dinov2}
Oquab, M., Darcet, T., Moutakanni, T., Vo, H., Szafraniec, M., Khalidov, V., Fernandez, P., Haziza, D., Massa, F., El-Nouby, A., et~al.: Dinov2: Learning robust visual features without supervision. arXiv preprint arXiv:2304.07193  (2023)

\bibitem{AngularGap2022}
Peng, B., Islam, M., Tu, M.: Angular gap: Reducing the uncertainty of image difficulty through model calibration. MM '22, Association for Computing Machinery, New York, NY, USA (2022)

\bibitem{dynamicPrune_2023}
Pentsos, V., Spantidi, O., Anagnostopoulos, I.: Dynamic image difficulty-aware \uppercase{DNN} pruning. Micromachines  \textbf{14}(5) (2023)

\bibitem{DLLagree2022}
Pliushch, I., Mundt, M., Lupp, N., Ramesh, V.: When deep classifiers agree: Analyzing correlations between learning order and image statistics. In: Avidan, S., et~al. (eds.) ECCV. pp. 397--413. Springer Nature Switzerland (2022)

\bibitem{dudaHart1973}
Richard O.~Duda, P.E.H.: Pattern Classification and Scene Analysis. Wiley-Interscience (1973)

\bibitem{appl_cancer}
Salman, M., Çakar, G., Azimjonov, J., Kösem, M., Cedi̇moğlu, I.: Automated prostate cancer grading and diagnosis system using deep learning-based yolo object detection algorithm. Expert Systems with Applications  \textbf{201},  117148 (2022)

\bibitem{HoughNet}
Samet, N., Hicsonmez, S., Akbas, E.: Houghnet: Integrating near and long-range evidence for bottom-up object detection. In: Eur. Conf. Comp. Vis. (ECCV) (2020)

\bibitem{sandler2018mobilenetv2}
Sandler, M., Howard, A., Zhu, M., Zhmoginov, A., Chen, L.C.: Mobilenetv2: Inverted residuals and linear bottlenecks. In: Proceedings of the IEEE conference on computer vision and pattern recognition. pp. 4510--4520 (2018)

\bibitem{DatasetClassification_2021}
Scheidegger, F., Istrate, R., Mariani, G., Benini, L., Bekas, C., Malossi, C.: Efficient image dataset classification difficulty estimation for predicting deep-learning accuracy. In: The Visual Computer. vol.~37, pp. 1593--1610 (2021)

\bibitem{schober2018correlation}
Schober, P., Boer, C., Schwarte, L.A.: Correlation coefficients: appropriate use and interpretation. Anesthesia \& analgesia  \textbf{126}(5),  1763--1768 (2018)

\bibitem{modelDiff2023}
Shah, H., Park, S.M., Ilyas, A., Madry, A.: {M}odel{D}iff: A framework for comparing learning algorithms. In: Proc. 40th Int. Conf. on Machine Learning. Proc.of Machine Learning Research, vol.~202, pp. 30646--30688 (23--29 Jul 2023)

\bibitem{vggNet2014}
Simonyan, K., Zisserman, A.: Very deep convolutional networks for large-scale image recognition. arXiv preprint arXiv:1409.1556  (2014)

\bibitem{Ionescu_2Stage_2018}
Soviany, P., Ionescu, R.T.: Optimizing the trade-off between single-stage and two-stage deep object detectors using image difficulty prediction. In: 2018 20th Int. Symposium on Symbolic and Numeric Algorithms for Scientific Computing (SYNASC). pp. 209--214 (2018)

\bibitem{food2018}
Subhi, M.A., Md.~Ali, S.: A deep convolutional neural network for food detection and recognition. In: 2018 IEEE-EMBS Conf. on Biomedical Engineering and Sciences (IECBES). pp. 284--287 (2018)

\bibitem{fish2018}
Tamou, A., Benzinou, A., Nasreddine, K., Ballihi, L.: Transfer learning with deep convolutional neural network for underwater live fish recognition. In: 2018 IEEE Int. Conf. on Image Processing, Appl.s and Systems (IPAS). pp. 204--209 (2018)

\bibitem{tan2019efficientnet}
Tan, M., Le, Q.: Efficientnet: Rethinking model scaling for convolutional neural networks. In: Int. Conf. on machine learning. pp. 6105--6114 (2019)

\bibitem{Ionescu_2016_CVPR}
Tudor~Ionescu, R., Alexe, B., Leordeanu, M., Popescu, M., Papadopoulos, D.P., Ferrari, V.: How hard can it be? estimating the difficulty of visual search in an image. In: Proc. IEEE Conf. Computer Vision and Pattern Recognition (June 2016)

\bibitem{pcc}
Wicklin, R.: Weak or strong? how to interpret a spearman or kendall correlation. \url{https://blogs.sas.com/content/iml/2023/04/05/interpret-spearman-kendall-corr.html} (2024), \url{https://blogs.sas.com/content/iml/2023/04/05/interpret-spearman-kendall-corr.html}, accessed on 2024-06-04

\bibitem{flowers2018}
Wu, Y., Qin, X., Pan, Y., Yuan, C.: Convolution neural network based transfer learning for classification of flowers. In: 2018 IEEE 3rd Int. Conf. on Signal and Image Processing (ICSIP). pp. 562--566 (2018)

\bibitem{appl_hardhat}
Y.~Li, H.~Wei, Z.H.J.H.W.W.: Deep learning-based safety helmet detection in engineering management based on convolutional neural networks. Advances in Civil Engineering  \textbf{2020},  88--97 (2020)

\bibitem{zhang2018shufflenet}
Zhang, X., Zhou, X., Lin, M., Sun, J.: Shufflenet: An extremely efficient convolutional neural network for mobile devices. In: Proc. IEEE Conf. on computer vision and pattern recognition. pp. 6848--6856 (2018)

\bibitem{zhou2023learnware}
Zhou, Z.H., Tan, Z.H.: Learnware: small models do big. Science China Information Sciences  \textbf{67},  1869--1919 (2023)

\bibitem{transferLearn2021}
Zhuang, F., Qi, Z., Duan, K., Xi, D., Zhu, Y., Zhu, H., Xiong, H., He, Q.: A comprehensive survey on transfer learning. Proc. IEEE  \textbf{109}(1),  43--76 (2021)

\end{thebibliography}
\bibliographystyle{splncs04}
\end{document}